\documentclass[10pt,twocolumn,letterpaper]{article}

\usepackage{iccv}
\usepackage{times}
\usepackage{epsfig}
\usepackage{graphicx}
\usepackage{amsmath}
\usepackage{amssymb}
\usepackage{verbatim}
\usepackage{subcaption}
\usepackage{multirow}
\usepackage{gensymb}
\usepackage{mathrsfs}
\usepackage{booktabs}
\usepackage{authblk}

\newcommand{\norm}[1]{\left\lVert#1\right\rVert}

\usepackage{color}

\usepackage[pagebackref=true,breaklinks=true,letterpaper=true,colorlinks,bookmarks=false]{hyperref}

\iccvfinalcopy % *** Uncomment this line for the final submission

\ificcvfinal\fi
\begin{document}

%%%%%%%%% TITLE
\title{Image-based Localization using Hourglass Networks}

%\author{Iaroslav Melekhov\\
%Aalto University\\
%{\tt\small iaroslav.melekhov@aalto.fi}
%\and
%Juha Ylioinas\\
%Aalto University\\
%{\tt\small juha.ylioinas@aalto.fi}
%\and
%Juho Kannala\\
%Aalto University\\
%{\tt\small juho.kannala@aalto.fi}
%\and
%Esa Rahtu\\
%University of Oulu\\
%{\tt\small esa.rahtu@ee.oulu.fi}
%}

\author[1]{Iaroslav Melekhov}
\author[1]{Juha Ylioinas}
\author[1]{Juho Kannala}
\author[2]{Esa Rahtu}
\affil[1]{Aalto University, Finland \authorcr
       {\tt\small firstname.lastname@aalto.fi}}
\affil[2]{Tampere University of Technology, Finland \authorcr
       {\tt\small esa.rahtu@tut.fi}}

\maketitle
%\thispagestyle{empty}

%%%%%%%%% ABSTRACT
\begin{abstract}
In this paper, we propose an encoder-decoder convolutional neural network (CNN) architecture for estimating camera pose (orientation and location) from a single RGB-image. The architecture has a hourglass shape consisting of a chain of convolution and up-convolution layers followed by a regression part. The up-convolution layers are introduced to preserve the fine-grained information of the input image. Following the common practice, we train our model in end-to-end manner utilizing transfer learning from large scale classification data. The experiments demonstrate the performance of the approach on data exhibiting different lighting conditions, reflections, and motion blur. The results indicate a clear improvement over the previous state-of-the-art even when compared to methods that utilize sequence of test frames instead of a single frame.

%An essential part of our work is an upconvolutional step which preserves fine-grained visual information of an input image vanishing in the later part of CNNs. We show that this is the key to more accurate camera pose estimates using a CNN-based approach applied to monocular images (i.e. without utilizing videos and/or applying LSTM units in the later steps of the pipeline). Following the previous studies, the proposed model is trained an end-to-end manner utilizing transfer learning from large scale classification data. We demonstrate the robustness and effectiveness of our approach by evaluating the network on data exhibiting different lightning variances, reflections and motion blur, which usually causes methods based on local hand-crafted feature descriptors to fail, such as SIFT. Our experimental results confirm that the proposed method outperforms previously reported state-of-the-art CNN-based approaches showing significant improvement in image-based localization.
%Source code and trained models for our experiments are publicly available at 

\end{abstract}

%%%%%%%%% BODY TEXT
\section{Introduction}
%\subsection{Contribution}
Image-based localization, or camera relocalization refers to the problem of estimating camera pose (orientation and position) from visual data. It plays a key role in many computer vision applications, such as simultaneous localization and mapping (SLAM), structure from motion (SfM), autonomous robot navigation, and augmented and mixed reality. Currently, there are plenty of relocalization methods proposed in the literature. However, many of these approaches are based on finding matches between local features extracted from an input image (by usually applying local image descriptor methods such as SIFT, ORB, or SURF~\cite{SIFT,ORB,SURF}) and features corresponding to 3D points in a model of the scene. In spite of their popularity, feature-based methods are not able to find matching points accurately in all scenarios. In particular, extremely large viewpoint changes, occlusions, repetitive structures and textureless scenes often produce simply too many outliers in the matching process. In order to cope with many outliers, the typical first aid is to apply RANSAC which unfortunately increases time and computational costs.

The increased computational power of graphic processing units (GPUs) and the availability of large-scale training datasets have made Convolutional Neural Networks (CNNs) the dominant paradigm in various computer vision problems, such as image retrieval~\cite{ImageRetrievalBabenko,DeepImageRetrieval}, object recognition, semantic segmentation, and image classification~\cite{ImagenetKrizh,ResNet}. For image-based localization, CNNs were considered for the first time by Kendall \etal~\cite{Posenet}. Their method, named PoseNet, casts camera relocalization as a regression problem, where 6-DoF camera pose is directly predicted from a monocular image by leveraging transfer learning from a large scale classification data. Although PoseNet overcomes many limitations of the feature-based approaches, its localization performance still lacks behind traditional approaches in typical cases where local features perform well.%for indoor image-based localization is poor relative to spatial extent of the scene.

Looking for possible ways to further improve the accuracy of image-based localization using CNN-based architectures, we adopt some recent advances discovered in efforts solving the problems of image restoration~\cite{RestorationMao}, semantic segmentation~\cite{SemanticSegmentationOliveira} and human pose estimation~\cite{HourglassNewell}. 
Inspired by these ideas, we propose to add more context to the regression process to better collect the overall information, from coarse structures to fine-grained object details, available in the input image. We argue that this kind of a mechanism is suitable for getting an accurate camera pose estimate using CNNs. In detail, we propose a network architecture which consists of a bottom part (the encoder) that is used to encode the overall context and a latter part (the decoder) that recovers the fine-grained visual information by up-convolving the output feature map of the encoder by gradually increasing its size towards the original resolution of the input image. Such a symmetric "encoder-decoder" network structure is also known as an hourglass architecture~\cite{HourglassNewell}.

% Contributions of this paper
The contributions of this paper can be summarized as follows:
\begin{itemize}
  \item We complement a deep convolutional network by adding a chain of up-convolutional layers with shortcut connections and apply it to the image-based localization problem. 
  %\item We show that the shortcuts connecting corresponding convolutional and upconvolutional layers allow to pass fine-grained visual information of input images from top to bottom layers of the network, and thus to predict camera orientation more accurately. Furthermore, as observed in~\cite{ResNet}, the shortcuts make the back-propagation of the gradients more fluent allowing to train deeper architectures easier.
  \item The proposed network significantly outperforms the current state-of-the-art methods proposed in the literature for estimating camera pose.
\end{itemize}

The remainder of this paper is organized as follows. Section~\ref{lbl:sec_related_work} discusses related work. In Section~\ref{lbl:sec_method} we provide the details of the proposed CNN architecture. Section~\ref{lbl:sec_experiments} presents the experimental methodology and results on a standard evaluation dataset. We conclude with a summary and ideas for future work.

The source code and trained models will be publicly available upon publication.

\begin{figure*}[t!]
\centering
\includegraphics[width=\textwidth]{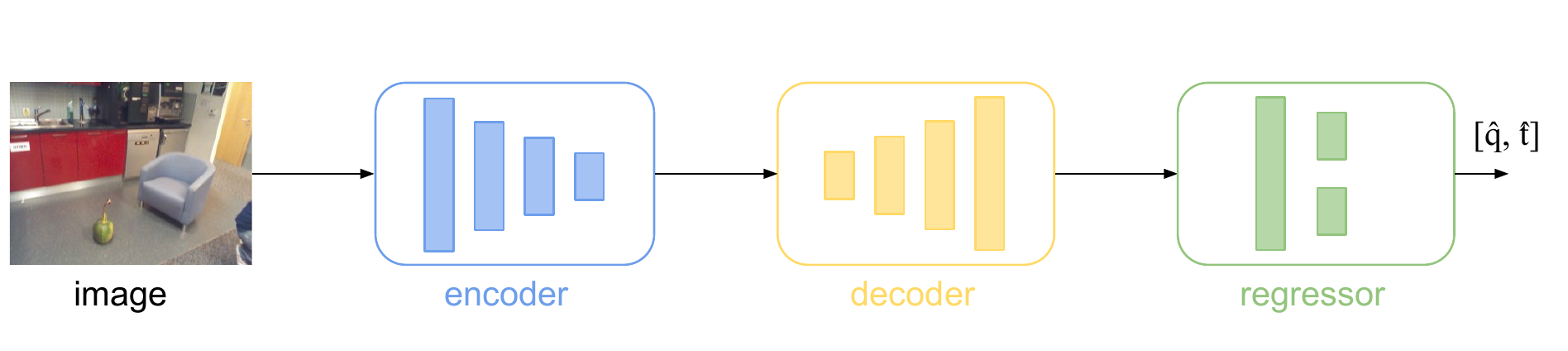}
\caption{Overview of our proposed architecture. It takes an RGB-image as input and predicts the camera pose. The overall network consists of three components, namely encoder, decoder and regressor. The encoder is fully convolutional up until a certain spatial resolution. The decoder then gradually increases the resolution of the feature map which is eventually fed to the regressor that is composed of three fully connected layers.}\label{lbl:fig_architecture_overview}
\end{figure*}

\section{Related Work}\label{lbl:sec_related_work}
Image-based localization can be solved by casting it as a place recognition problem. In this approach, image retrieval techniques are often applied to find similar views of the scene in a database of images for which camera position is known. The method then estimates an approximate camera pose using the information in retrieved images. As noted in \cite{SCOREver2}, these methods suffer in situations where there are no strong constraints on the camera motion. This is due to the number of the key-frames that is often very sparse. 

Perhaps a more traditional approach to image-based localization is based on finding correspondences between a query image and a 3D scene model reconstructed using SfM. Given a query image and a 3D model, an essential part of this approach is matching points from 2D to 3D. The main limitation of this approach is the 3D model that may grow eventually too big in its size or just go too complex if the scene itself is somehow complicated, like large-scale urban environments. In such scenarios, the ratio of outliers in the matching process often grows too high. This in turn results in a growth in the run-time of RANSAC. There are methods to handle this situation, such as prioritizing matching regions in 2D to 3D and/or 3D to 2D and using co-visibility of the query and the model \cite{ActiveSearch}. 

Applying machine learning techniques has proven very effective in image-based indoor localization. Shotton \etal~\cite{SCORE} proposed a method to estimate scene coordinates from an RGB-D input using decision forests. Compared to traditional algorithms based on matching point correspondences, their method removes the need for the traditional pipeline of feature extraction, feature description, and matching.  Valentin \etal \cite{SCOREver2} further improved the  method by exploiting  uncertainty in the model in order to move from sole point estimates to predict also their uncertainties for more robust continuous pose optimization.  Both of these methods are designed for cameras that have an RGB-D sensor.

Very recently, applying deep learning techniques has resulted in remarkable performance improvements in many computer vision problems~\cite{ImageRetrievalBabenko, RestorationMao, SemanticSegmentationOliveira}. Partly motivated by studies applying CNNs and regression~\cite{GoogLeNet,Saliency,PoseEstimation}, Kendall~\etal~\cite{Posenet} proposed an architecture trying to directly regress camera relocalization from an input RGB image. More recent CNN-based approaches cover those of Clark~\etal~\cite{VidLoc} and Walch~\etal~\cite{LSTMPose}. Both of these follow~\cite{Posenet}, and similarly adopt the same CNN architecture, by pre-training it first on large-scale image classification data, for extracting features from input images to be localized. In detail, Walch~\etal~\cite{LSTMPose} consider these features as an input sequence to a block of four LSTM units operating along four directions (up, down, left, and right) independently. On top of that, there is a regression part which encompasses fully-connected layers for predicting the camera pose. In turn, Clark~\etal~\cite{VidLoc} applied LSTMs to predict camera translation only, but using short videos as an input. Their method is a bidirectional recurrent neural network (RNN), which captures dependencies between adjacent image frames yielding refined accuracy of the global pose. Both of the two architectures lead to improvement in the accuracy of 6-DoF camera pose outperforming PoseNet~\cite{Posenet}.

Compared to non-CNN based approaches, our method belongs to the very recent initiative of models that do not require any online 3D models in camera pose estimation. In contrast to \cite{SCORE,SCOREver2}, our method is solely based on monocular RGB images and no depth information is required. Compared to PoseNet \cite{Posenet}, our method aims at better utilization of context and provides improvement in pose estimation accuracy. In comparison to \cite{LSTMPose}, our method is more accurate in indoor locations. Finally, our method does not rely on video inputs, but still outperforms the CNN-model presented in \cite{VidLoc} for video-clip relocalization.

%In contrast, our approach and~\cite{LSTMPose,Posenet} methods performing localization of monocular images. Moreover, VidLoc recovers only position and not orientation which is essential to Structure-from-Motion.

%The detailed evaluation of some architectural variations of our approach are presented in Section~\ref{lbl:sec_experiments}.

\begin{figure*}[t!]
\centering
\includegraphics[width=.135\textwidth]{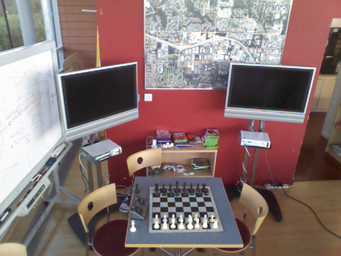}
\includegraphics[width=.135\textwidth]{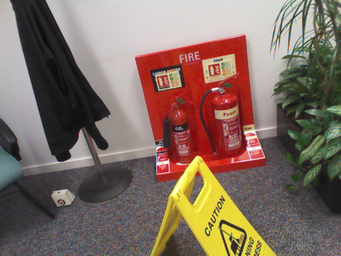}
\includegraphics[width=.135\textwidth]{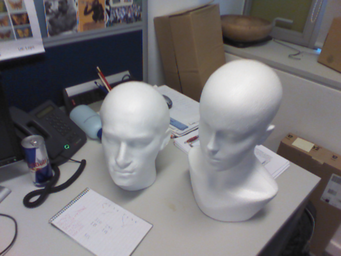}
\includegraphics[width=.135\textwidth]{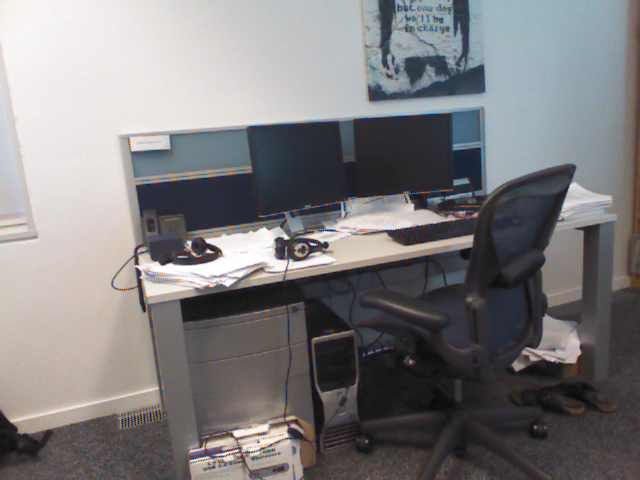}
\includegraphics[width=.135\textwidth]{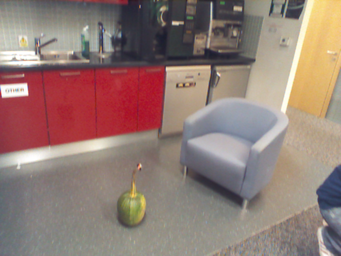}
\includegraphics[width=.135\textwidth]{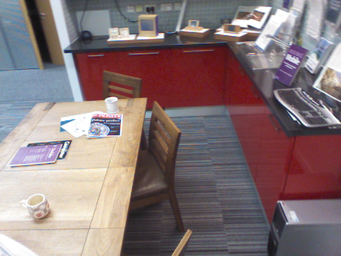}
\includegraphics[width=.135\textwidth]{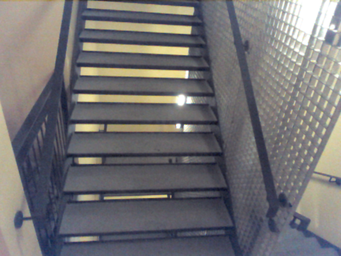}
\caption{Visual representation of the categories of 7-Scenes dataset. From left to right: Chess, Fire, Heads, Office, Pumpkin, Red Kitchen and Stairs.}\label{lbl:fig_7scenes_examples}
\end{figure*}

\section{Method}\label{lbl:sec_method}
Following \cite{Posenet,LSTMPose}, our goal is to estimate camera pose directly from an RGB image. We propose a CNN architecture that predicts a 7-dimensional camera pose vector $\mathbf{p}\!=\!\left[\mathbf{q},\mathbf{t}\right]$ consisting of an orientation component $\mathbf{q}=\left[q_{1},q_{2},q_{3},q_{4}\right]$ represented by quaternions and a translation component $\mathbf{t}=\left[t_{1},t_{2},t_{3}\right]$.

Hiding the architectural details, the overall network structure is illustrated in Fig.~\ref{lbl:fig_architecture_overview}. The network consists of three components, namely \emph{encoder}, \emph{decoder} and \emph{regressor}. The encoder is fully convolutional acting as a feature extractor. The decoder consists of up-convolutional layers stacked to recover the fine-grained details of the input from the decoder outputs. Finally, the decoder is followed by the regressor that estimates the camera pose $\mathbf{p}$.
%. In this section 

%A pair of images is fed to a regression neural network consisting of two branches which directly estimates the real-valued parameters of the relative camera pose vector using the ground truth information
%\subsection{Network Loss}
To train our hourglass-shaped CNN model, we apply the following objective function~\cite{Posenet}:

\begin{equation}
\mathscr{L}=\norm{\mathbf{t} - \hat{\mathbf{t}}} + \beta \norm{\mathbf{q} - \frac{\hat{\mathbf{q}}}{\norm{\hat{\mathbf{q}}}}},\label{lbl:eq_loss_function}
\end{equation}
where $(\mathbf{t},\mathbf{q})$ and $(\hat{\mathbf{t}},\hat{\mathbf{q}})$ are ground truth and estimated
translation-orientation pairs, respectively. $\beta$ is a scale factor, tunable by grid search, that keeps the estimated orientation and translation to be nearly equal. The quaternion based orientation vector $\mathbf{q}$ is normalized to unit length at test time. We provide the detailed information about the other hyperparameters used in training in Section~\ref{lbl:sec_experiments}.
%Following the original work~\cite{Posenet}, 

\subsection{CNN Architecture}
Training convolutional neural networks from scratch for image-based localization task is impractical due to the lack of training data. Following \cite{Posenet}, we leverage a pre-trained large-scale classification network. Specifically, to find a balance between the number of parameters of the network and accuracy, we adopt ResNet34~\cite{ResNet} architecture which has good performance among other classification approaches~\cite{CNNComparison} as our base network. We remove the last fully-connected layer from the original ResNet34 model but keep the convolutional and pooling layers intact. The resulting architecture is considered as the encoder part of the whole pipeline.

Instead of connecting the encoder to the regression part directly, we propose to add some extra layers between them. In detail, we add three up-convolutional and one convolutional layer. The main idea of using up-convolutional layers is to restore essential fine-grained visual information of the input image lost in encoder part of the network. Up-convolutional layers have been widely applied in image restoration~\cite{RestorationMao}, structure from motion~\cite{Demon} and semantic segmentation~\cite{SegmentationHong,SegmentationNoh}. The proposed architecture is presented in Fig.~\ref{lbl:fig_hourglass_detailed}.

Finally, there is a regressor module on top of the encoder. The regressor consists of three fully connected layers, namely localization layer, orientation layer and translation layer. In contrast to the regressor originally proposed in~\cite{Posenet}, we slightly modified its architecture by appending batch-normalization after each fully connected layer. 

Inspired by the visualization of the steps of downsampling and upsampling of the feature maps flowing through encoder-decoder part and by~\cite{HourglassNewell}'s work, we call our CNN architecture \texttt{Hourglass-Pose}.

%\begin{figure}[h!]
%\centering
%\includegraphics[width=.5\textwidth]{figures/resnet_pose_architecture}
%\caption{ResNet34-Pose architecture for estimating camera pose $\mathbf{P}=\left[\mathbf{q},\mathbf{t}%\right]$ by processing a single RGB-image. The image goes through an encoder represented by pre-trained %ResNet34 model. The output of the encoder is fed to a regressor consisting of a set of FC-layers to %predict $\mathbf{q}$ and $\mathbf{t}$ respectively. The number of connections of each FC-layer is given %in parenthesis.}\label{lbl:fig_resnet_pose}
%\end{figure}

\subsubsection{Hourglass-Pose}
As explained, the encoder part of our architecture is the slightly modified ResNet34 model. It differs from the original one presented in \cite{ResNet} so that the final softmax layer and the last average pooling layer have been removed. As a result the spatial resolution of the encoder feature map is $7\times7$. 

To better preserve finer details of the input image for the localization task, we added skip (shortcut) connections from each of the four residual blocks of the encoder to the corresponding up-convolution and the final convolution layers of the decoder. The last part of the decoder, namely the final convolutional module (a chain of convolutional, batch-normalization~\cite{BatchNorm} and ReLU layers) does not alter the spatial resolution of the feature map ($56\times56$), but is used to decrease the number of channels. In our preliminary experiments, we also experimented with a Spatial Pyramid Pooling (SPP) layer~\cite{SPP} instead of the convolutional module. Particularly, SPP layer consists of a set of pooling layers (pyramid levels) producing a fix-sized feature map regardless the size of the input image. However, the camera pose estimations were not improved, and we omitted SPP in favor of simpler convolutional module. The encoder-decoder module is followed by a regressor which predicts the camera orientation $\textbf{q}$ and translation $\textbf{t}$. The detailed network configuration is shown in Table~\ref{lbl:tbl_hourglass_pose_structure}.

\begin{figure*}[t!]
\centering
\includegraphics[width=\textwidth]{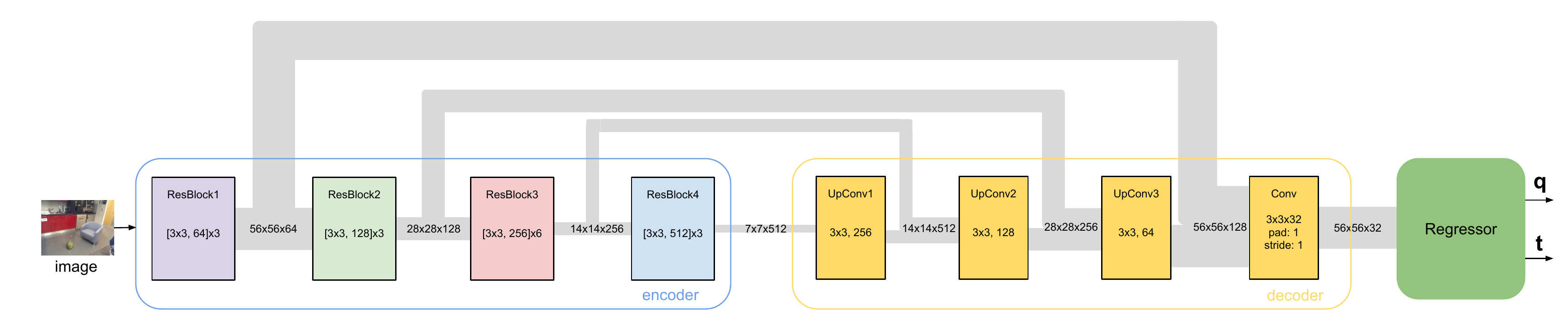}
\caption{An illustration of the proposed architecture referred to as Hourglass-Pose for predicting camera pose. The encoder part is a modified version of ResNet34~\cite{ResNet}, where % pre-trained on ImageNet~\cite{ImageNet}. 
 We removed the last fully-connected and average pooling layers from the original ResNet34 arhitecture and kept only the convolutional layers. The decoder consists of a set of stacked up-convolutional layers gradually increasing the spatial resolution of the feature maps up to $56\times56$. We further added one convolutional layer for dimensionality reduction. Skip connections connect each block of the encoder to the corresponding parts of the decoder allowing the decoder to re-utilize features from the earlier layers of the network. Finally, camera pose is estimated by the regressor as explained in Section 3.}\label{lbl:fig_hourglass_detailed}
\end{figure*}

\begin{figure}[t!]
\centering
\includegraphics[width=0.5\textwidth]{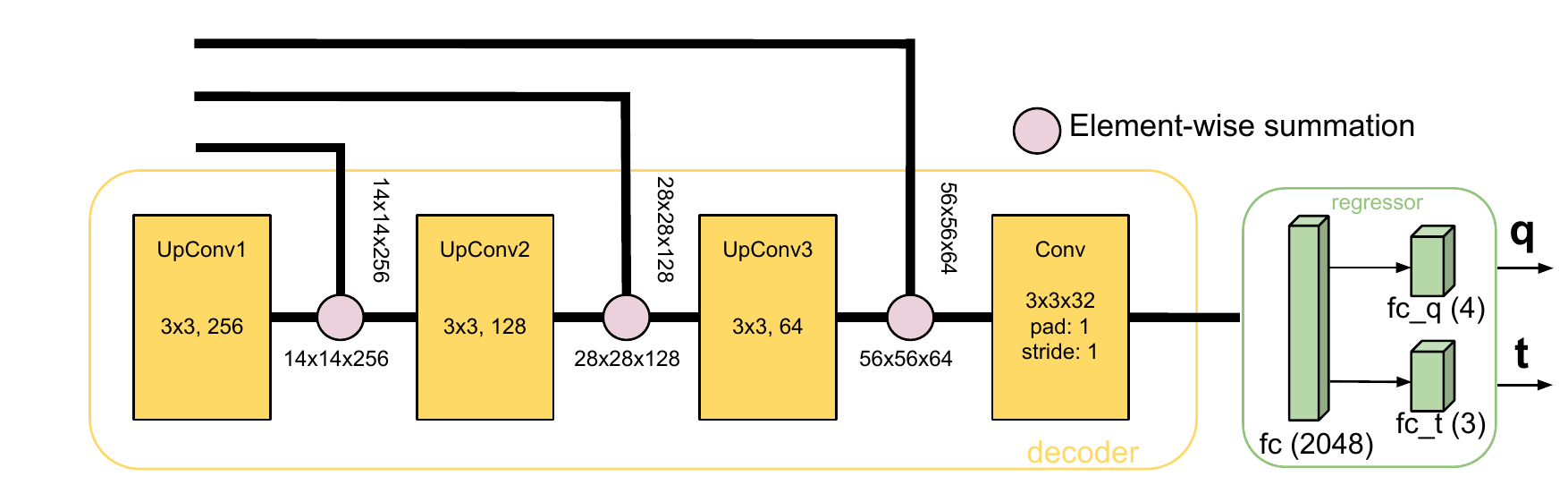}
\caption{The structure of the decoder and the regressor of HourglassSum-Pose architecture for estimating camera pose $\mathbf{p}=\left[\mathbf{q},\mathbf{t}\right]$. The output of the decoder is connected to the regressor consisting of a set of FC-layers to predict $\mathbf{q}$ and $\mathbf{t}$ respectively. The number of connections of each FC-layer is given in parenthesis.}\label{lbl:fig_hourglasssum_architecture}
\end{figure}

In order to investigate the benefits of using skip connections more thoroughly, we experimented with different aggregation strategies of the encoder and the decoder feature maps. In contrast to \texttt{Hourglass-Pose} where the outputs of corresponding layers are concatenated (See Fig.~\ref{lbl:fig_hourglass_detailed}), we evaluated the whole pipeline by also calculating an element-wise sum of the feature maps connected via skip connections. We refer to the corresponding architecture as \texttt{HourglassSum-Pose}. Schematic illustration of a decoder-regressor part of this structure is presented in Fig.~\ref{lbl:fig_hourglasssum_architecture}. %Finally, we found that the elementwise sum yields the best performance.

%In our case, we want to pass information of the convolutional feature maps to the corresponding upconvolutional layers
%~\cite{StrivingForSimplicity}

\begin{table}
\begin{center}
\resizebox{.5\textwidth}{!}{%
	\begin{tabular}{c  c  c  c}
	\hline
	Module & Layers & Output Size & Hourglass-Pose \\ \hline

	\parbox[c]{4mm}{\multirow{10}{*}{\rotatebox[origin=c]{90}{Encoder}}}    & Conv & $112\times112$ & $7\times7$, $64$  (3 / s2) \\  
	       & Pool     & $56\times 56$  & $3\times3$ max (s2)    \\   \addlinespace[1ex]
	       & ResBlock1   & $56\times 56$   & \small $\left[ \begin{array}{cc}
	       												3\times3, 64 \\
	       												3\times3, 64
											       \end{array} \right]\times3$ (1 / s1)  \\ 
\addlinespace[1ex]
		   & ResBlock2   & $28\times28$   & \small $\left[ \begin{array}{cc}
	       												3\times3, 128 \\
	       												3\times3, 128
											       \end{array} \right]\times4$ (1 / s1)  \\ 
\addlinespace[1ex]
           & ResBlock3   & $14\times14$   & \small $\left[ \begin{array}{cc}
	       												3\times3, 256 \\
	       												3\times3, 256
											       \end{array} \right]\times6$ (1 / s1)  \\ 
\addlinespace[1ex]
		   & ResBlock4   & $7\times7$   & \small $\left[ \begin{array}{cc}
	       												3\times3, 512 \\
	       												3\times3, 512
											       \end{array} \right]\times3$ (1 / s1)  \\ \addlinespace[1ex] \hline
\addlinespace[1ex]
\parbox[c]{4mm}{\multirow{4}{*}{\rotatebox[origin=c]{90}{Decoder}}} & UpConv1 & $14\times14$ & $4\times4$ upconv (1 / s2) \\
           & UpConv2 & $28\times28$ & $4\times4$ upconv (1 / s2) \\
           & UpConv3 & $56\times56$ & $4\times4$ upconv (1 / s2) \\
           & Conv & $56\times56$ & $3\times3$, 32, (1 / s1) \\   \hline
\addlinespace[1ex]
\parbox[c]{4mm}{\multirow{3}{*}{\rotatebox[origin=c]{90}{Regressor}}}   & FC & 2048 &  \\
           & FC$_{\mathbf{q}}$   & 4 & fully-connected \\
           & FC$_{\mathbf{t}}$   & 3 &  \\ \addlinespace[1ex] \hline
	\end{tabular}
	}
\end{center}
\caption{Details of our Hourglass-Pose architecture for estimating camera pose. Note that each convolutional/upconvolutional layer in the encoder-decoder part corresponds 'Conv-ReLU-BatchNorm' sequence. In Resblocks the resolution is downsampled with stride 2 convolutions. Upconvolution is implemented by first upsampling the signal by zero padding and then by applying normal convolution.}\label{lbl:tbl_hourglass_pose_structure}
\end{table}

\subsection{Evaluation Dataset}
To evaluate our method and compare with the state-of-the-art approaches, we utilize Microsoft 7-Scenes Dataset containing RGB-D images of 7 different indoor locations~\cite{7ScenesDataset}. The dataset has been widely used for camera relocalization~\cite{realTimeRelocalization,Posenet,LSTMPose,VidLoc}. The images of the scenes were recorded with a camera of the Kinect device at $640 \times 480$ resolution and divided to train and evaluation parts accordingly. The ground truth camera poses were obtained by applying the KinectFusion algorithm~\cite{KinnectFusion} producing smooth camera trajectories. Sample images covering all scenes of the dataset are illustrated in Fig.~\ref{lbl:fig_7scenes_examples}. They represent indoor views of the 7 scenes exhibiting different lighting conditions, textureless (\eg two statues in 'Heads') and repeated objects ('Stairs' scene), changes in viewpoint and motion blur. All of these factors make camera pose estimation an extremely challenging problem.

\section{Experiments}\label{lbl:sec_experiments}
In the following section we empirically demonstrate the effectiveness of the proposed approach on the 7-Scenes evaluation dataset and compare it to other state-of-the-art CNN-based methods. Like it was done in~\cite{Posenet}, we report the median error of camera orientation and translation in our evaluations.

\subsection{Other state-of-the-art approaches}
In this work we consider three recently proposed 6-DoF camera relocalization systems based on CNNs. 

\textbf{PoseNet} is \cite{Posenet} is based on the GoogLeNet~\cite{GoogLeNet} architecture. It processes RGB-images and is modified so that all three softmax and fully connected layers are removed from the original model and replaced by regressors in the training phase. In the testing phase the other two regressors of the lower layers are removed and the prediction is done solely based on the regressor on the top of the whole network.

\textbf{Bayesian PoseNet} Kendall~\etal~\cite{BayesianPosenet} propose a Bayesian convolutional neural network to estimate uncertainty in the global camera pose which leads to improving localization accuracy. The Bayesian convolutional neural is based on PoseNet architecture by adding dropout after the fully connected layers in the pose regressor and after one of the inception layer (layer 9) of GoogLeNet architecture.

%\paragraph{LSTM-Pose}
\textbf{LSTM-Pose}~\cite{LSTMPose} is otherwise similar to PoseNet, but applies LSTM networks for output feature coming from the final fully connected layer. In detail, it is based on utilizing the pre-trained GoogLeNet architecture as a feature extractor followed by four LSTM units applying in the up, down, left and right directions. The outputs of LSTM units are then concatenated and fed to a regression module consisting of two fully connected layers to predict camera pose.

%\paragraph{VidLoc}
\textbf{VidLoc}~\cite{VidLoc} is a CNN-based system based on short video clips. As in PoseNet and LSTM-Pose, VidLoc incorporates similarly modified pre-trained GoogLeNet model for feature extraction. The output of this module is passed to bidirectional LSTM units predicting the poses for each frame in the sequence by exploiting contextual information in past and future frames.
\def\Plus{\texttt{+}}

\begin{table*}[t!]
\begin{center}
\resizebox{1\textwidth}{!}{%
	\begin{tabular}{l l l l | l l l l l l}
    \hline
	\multirow{2}{*}{Scene} & \multicolumn{2}{c}{Frames} & \multicolumn{1}{c|}{Spatial} & \multicolumn{1}{c}{PoseNet}  & \multicolumn{1}{c}{Bayesian} & \multicolumn{1}{c}{LSTM-Pose} & \multicolumn{1}{c}{VidLoc} & \multirow{2}{*}{Hourglass-Pose} & \multirow{2}{*}{HourglassSum-Pose} \\
    & \multicolumn{1}{c}{Train} & \multicolumn{1}{c}{Test} & \multicolumn{1}{c|}{Extent} & \multicolumn{1}{c}{ICCV'15~\cite{Posenet}} & PoseNet~\cite{BayesianPosenet} & \multicolumn{1}{c}{\cite{LSTMPose}} & \multicolumn{1}{c}{\cite{VidLoc}} & & \\
    \hline
    Chess & 4000 & 2000 & $3\times 2 \times 1$m & $0.32$m, $8.12^{\circ}$ & $0.37$m, $7.24^{\circ}$ & $0.24$m, $5.77^{\circ}$ & $0.18$m, N/A & $0.15$m, $6.53^{\circ}$ & $0.15$m, $6.17^{\circ}$ \\
    Fire & 2000 & 2000 & $2.5\times 1 \times 1$m & $0.47$m, $14.4^{\circ}$ & $0.43$m, $13.7^{\circ}$ & $0.34$m, $11.9^{\circ}$ & $0.26$m, N/A & $0.29$m, $11.59^{\circ}$ & $0.27$m, $10.84^{\circ}$ \\
    Heads & 1000 & 1000 & $2\times 0.5 \times 1$m & $0.29$m, $12.0^{\circ}$ & $0.31$m, $12.0^{\circ}$ & $0.21$m,  $13.7^{\circ}$ & $0.14$m, N/A & $0.21$m, $14.52^{\circ}$ & $0.19$m, $11.63^{\circ}$ \\
    Office & 6000 & 4000 & $2.5\times 2 \times 1.5$m & $0.48$m, $7.68^{\circ}$ & $0.48$m, $8.04^{\circ}$ & $0.30$m, $8.08^{\circ}$ & $0.26$m, N/A & $0.21$m, $9.25^{\circ}$ & $0.21$m, $8.48^{\circ}$  \\
    Pumpkin & 4000 & 2000 & $2.5\times 2 \times 1$m & $0.47$m, $8.42^{\circ}$ & $0.61$m, $7.08^{\circ}$ & $0.33$m, $7.00^{\circ}$ & $0.36$m, N/A & $0.27$m, $6.93^{\circ}$ & $0.25$m, $7.01^{\circ}$  \\
    Red Kitchen & 7000 & 5000 & $4\times 3 \times 1.5$m & $0.59$m, $8.64^{\circ}$ & $0.58$m, $7.54^{\circ}$ & $0.37$m, $8.83^{\circ}$ & $0.31$m, N/A & $0.27$m, $9.82^{\circ}$ & $0.27$m, $10.15^{\circ}$ \\
    Stairs & 2000 & 1000 & $2.5\times 2 \times 1.5$m & $0.47$m, $13.8^{\circ}$ & $0.48$m, $13.1^{\circ}$ & $0.40$m, $13.7^{\circ}$ & $0.26$m, N/A & $0.29$m, $13.07^{\circ}$ & $0.29$m, $12.46^{\circ}$ \\
    \hline \hline
    Average & & & & $0.44$m, $10.4^{\circ}$ & $0.47$m, $9.81^{\circ}$ & $0.31$m, $9.85^{\circ}$ & $0.25$m, N/A & $0.24$m, $10.24^{\circ}$ & $0.23$m, $9.53^{\circ}$
	\end{tabular}
    }
\end{center}
\caption{Performance comparison of two architectures (Hourglass-Pose and HourglassSum-Pose) and state-of-the-art methods on 7-Scenes evaluation dataset. Numbers are median translation and orientation errors for the entire test subset of each scene. Both models significantly outperform PoseNet~\cite{ResNet} and LSTM-Pose~\cite{LSTMPose} in terms of localization. It is a crucial observation emphasizing the importance of re-utilizing feature maps by using direct (skip) connections between encoder and decoder modules for image-based relocalization task. An Hourglass-Pose and HourglassSum-Pose architectures' comparison reveals that applying element-wise summation is more beneficial than features concatenation providing more accurate camera pose. Remarkably, the proposed models do perform even better than VidLoc~\cite{VidLoc} approach, which uses a sequence of test frames to estimate camera pose.}\label{lbl:table_final_results}
\end{table*}

\subsection{Training Setup}
We trained our models for each scene of 7-Scenes dataset according to the data splits provided by~\cite{7ScenesDataset}.

For all of our methods, we take the weights of ResNet34~\cite{ResNet} pre-trained on ImageNet to initialize the encoder part with them. The weights of the decoder and the regressor are initialized according to~\cite{Xavier}. Our initial learning rate is $10^{-3}$ and that is kept for the first 50 epochs. Then, we continue for 40 epochs with $10^{-4}$ and subsequently decrease it to $10^{-5}$ for the last 30 epochs.
%We take this trained model as a starting point for our hourglass family architectures but initialize the weights of a decoder-regression part with Xavier~\cite{Xavier} random variables. 
%For the hourglass models we set the learning rate at $10^{-3}$ during the first 50 epochs, gradually decreasing to $10^{-4}$ and $10^{-5}$ after 50th and 70th epoch respectively. The training was stopped after 90 epochs.

As a preprocessing step, all images of the evaluation dataset are rescaled so that the smaller side of the image is always 256 pixels. We calculate mean and standard deviation of pixel intensities separately for each scene and use them to normalize intensity value of every pixel in the input image.

We trained our models using random crops ($224\times224$) and performed the evaluation using central crops at the test time. All experiments were conducted on two NVIDIA Titan X GPUs with data parallelism using Torch7~\cite{Torch7}. We minimize the loss function (\ref{lbl:eq_loss_function}) over a training part of each scene of the evaluation dataset using Adam~\cite{Adam} ($\beta_1=0.9$, $\beta_2=0.99$). The scale factor $\beta$ (\ref{lbl:eq_loss_function}) varies between $1$ to $10$. Training mini-batches are randomly shuffled in the beginning of each training epoch. We further used set the weight decay as $10^{-5}$, used a mini-batch size of $40$ and the dropout probability as $0.5$. These parameter values were kept fixed during our experiments.

\subsection{Results}
To compare Hourglass-Pose and HourglassSum-Pose architectures with other state-of-the-art methods, we follow the evaluation protocol presented in ~\cite{Posenet}. Specifically, we report the median error of camera pose estimations for all scenes of the 7-Scenes dataset. Like in \cite{BayesianPosenet,LSTMPose,VidLoc}, we also provide an average median orientation and translation error.

Table~\ref{lbl:table_final_results} shows the performance of our approaches along with the other state-of-the-art. The values for other methods are taken from~\cite{Posenet},~\cite{BayesianPosenet},~\cite{LSTMPose}, and~\cite{VidLoc}. According to the results, several conclusions can be drawn. First, our architectures clearly outperform the other state-of-the-art CNN-based approaches. In general, HourglassSum-Pose improves the accuracy of the camera position by 52.27\% and orientation by 8.47\% for average error with respect to PoseNet. Furthermore, HourglassSum-Pose manages to achieve better orientation accuracy than LSTM-Pose~\cite{LSTMPose} in all scenes of the evaluation dataset. It can be seen that both of our architectures are even competitive with VidLoc~\cite{VidLoc} that is based on a sequence of frames. Our methods improve the average position error by 1 cm and 2 cm. The results in Table~\ref{lbl:table_final_results} confirm that it is beneficial to utilize an hourglass architecture for image-based localization.

\begin{figure*}[t!]
	\centering
	\begin{subfigure}[t]{.5\textwidth}
		\centering
		\includegraphics[width=.5\textwidth]{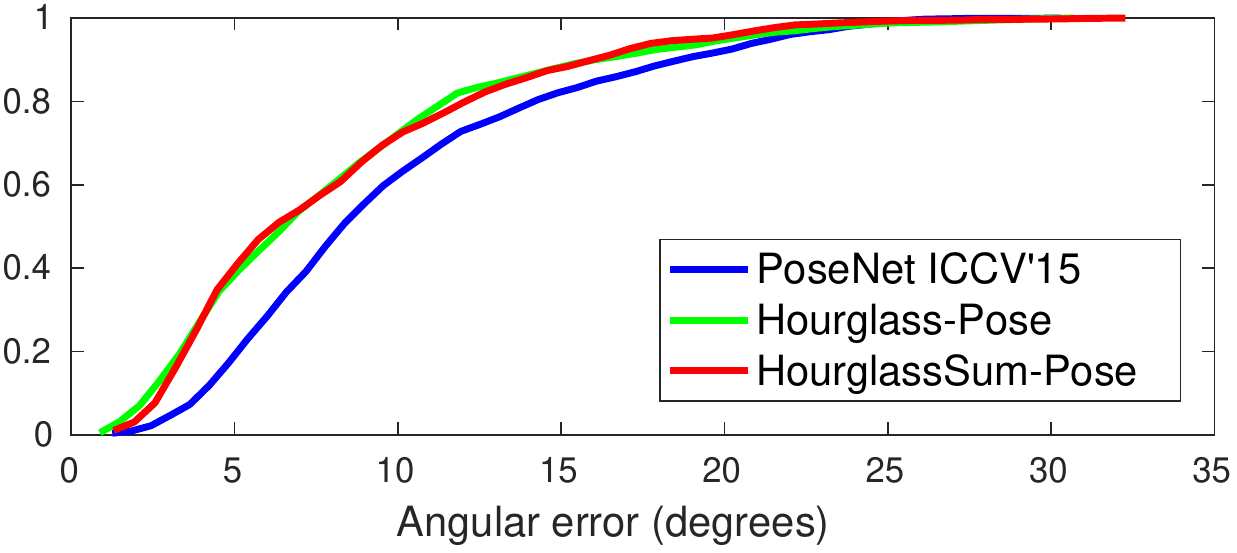}%
		\includegraphics[width=.5\textwidth]{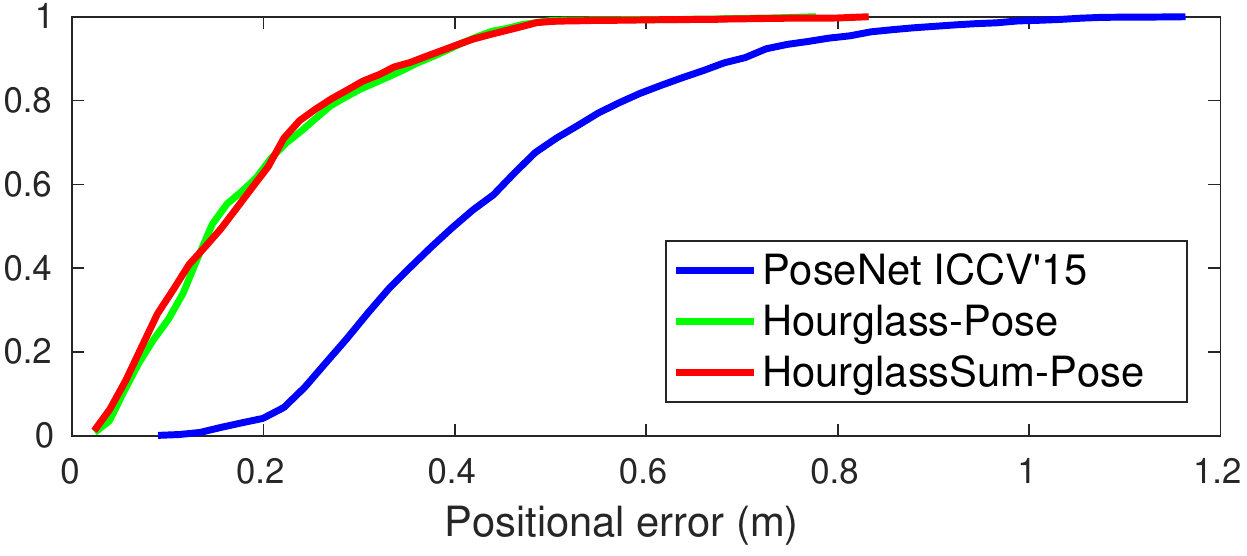}
		\caption{Chess}
	\end{subfigure}%
	~
	\begin{subfigure}[t]{.5\textwidth}
		\centering
		\includegraphics[width=.5\textwidth]{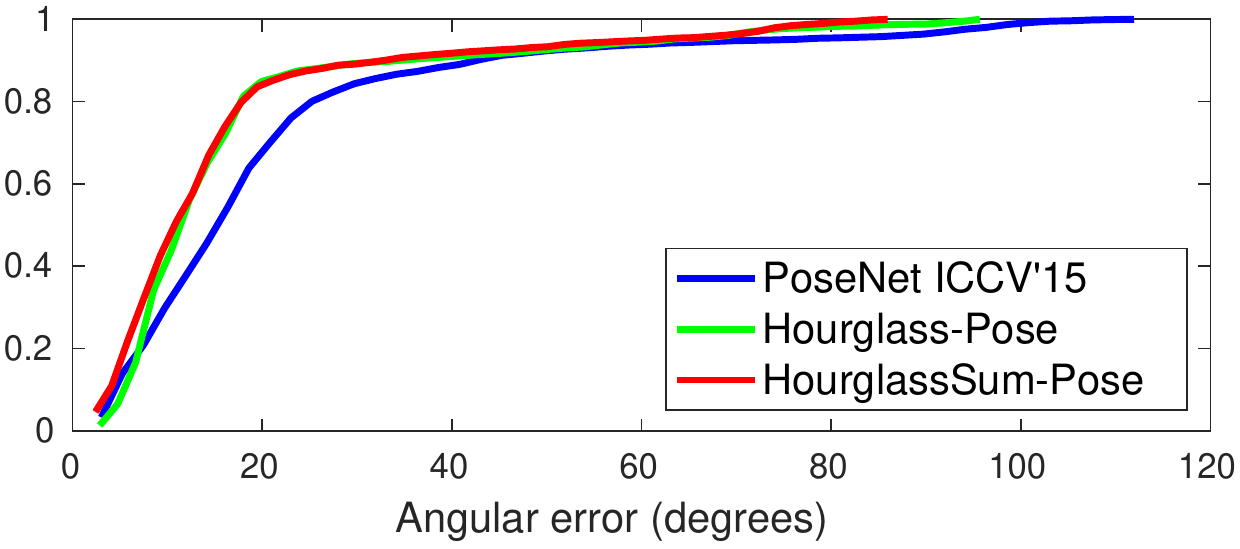}%
		\includegraphics[width=.5\textwidth]{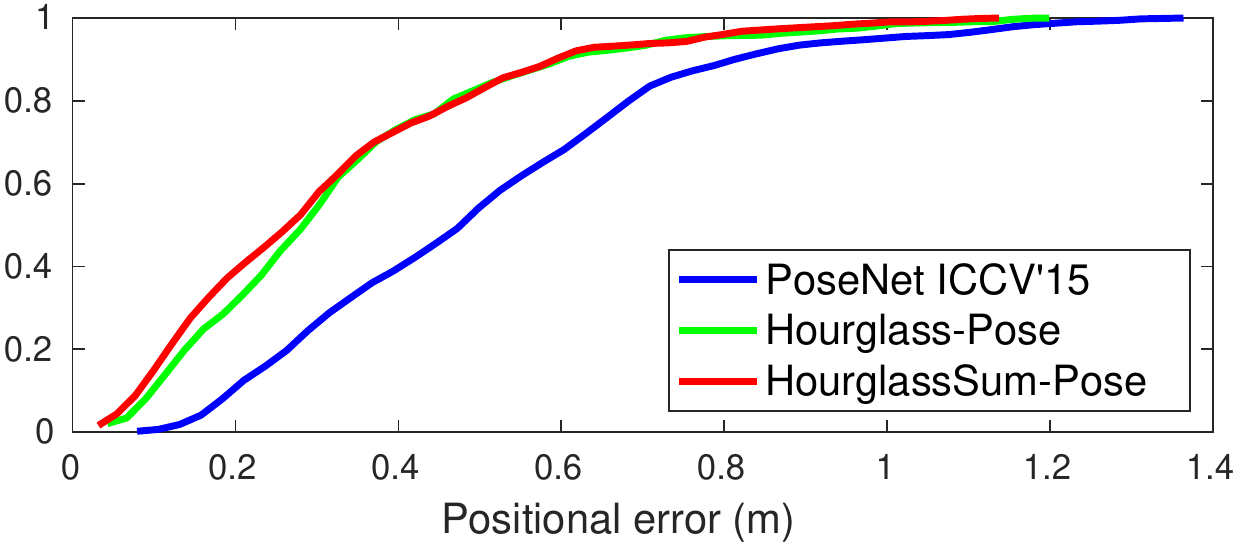}
		\caption{Fire}
	\end{subfigure}
	~
	\begin{subfigure}[t]{.5\textwidth}
		\centering
		\includegraphics[width=.5\textwidth]{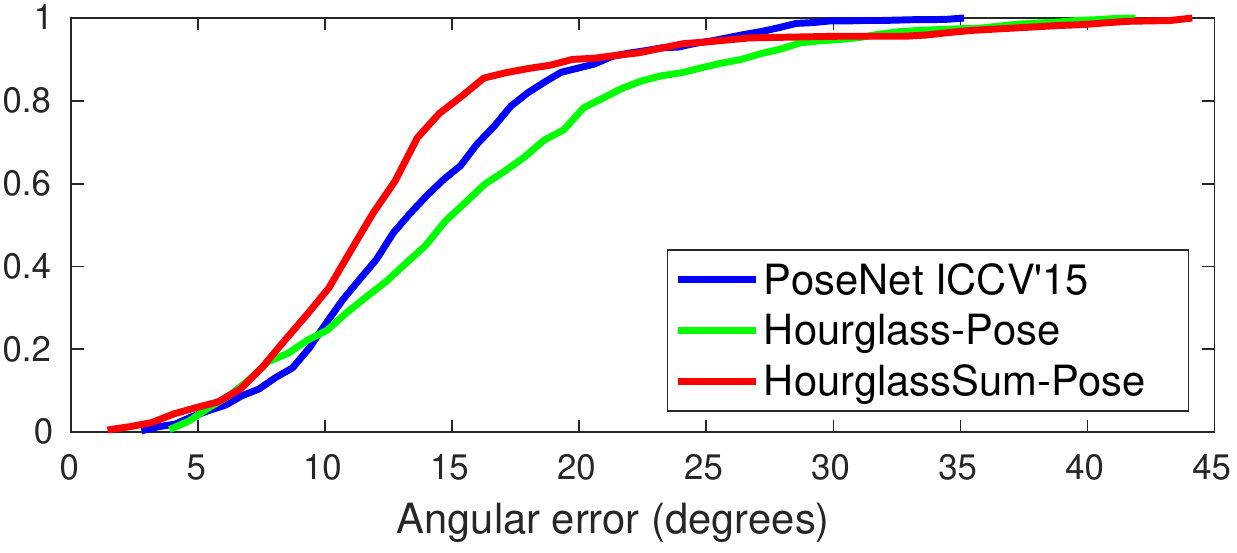}%
		\includegraphics[width=.5\textwidth]{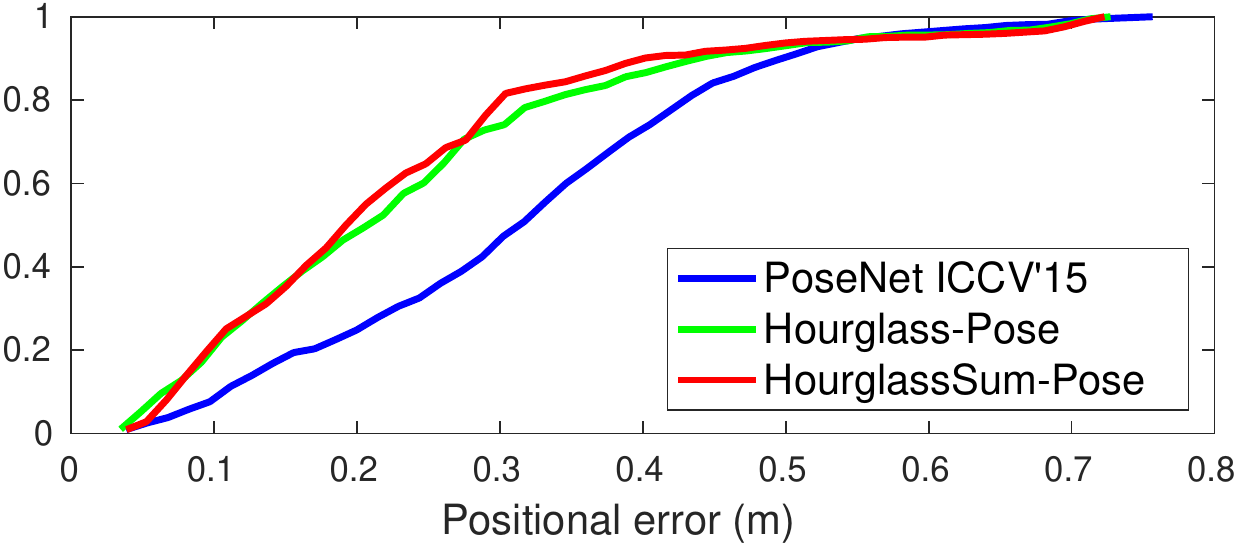}
		\caption{Heads}
	\end{subfigure}%
	~
	\begin{subfigure}[t]{.5\textwidth}
		\centering
		\includegraphics[width=.5\textwidth]{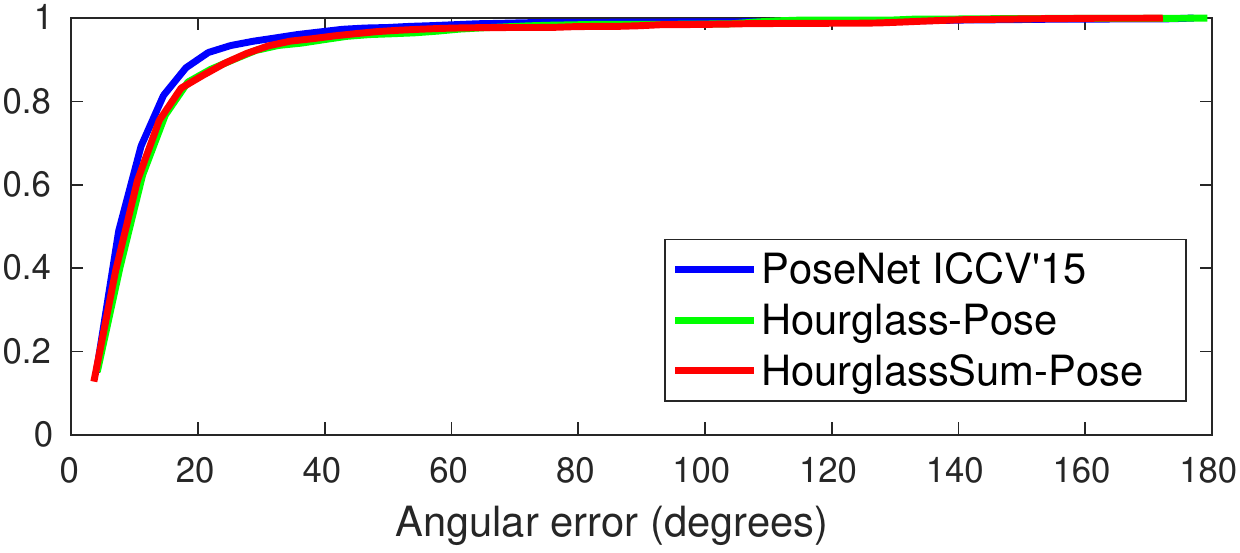}%
		\includegraphics[width=.5\textwidth]{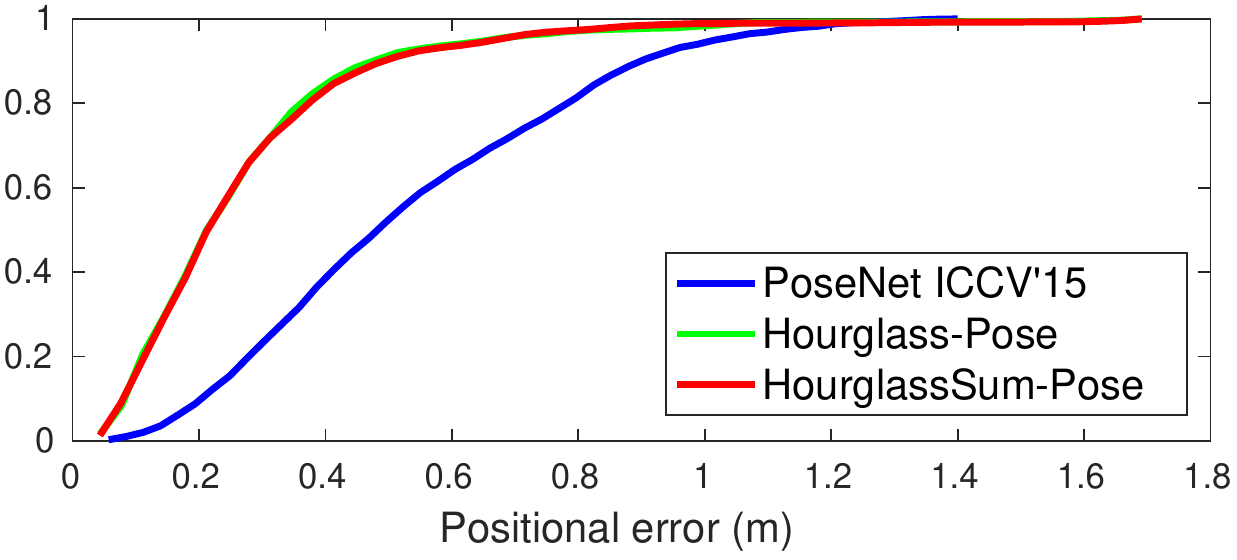}
		\caption{Office}
	\end{subfigure}
	~
	\begin{subfigure}[t]{.5\textwidth}
		\centering
		\includegraphics[width=.5\textwidth]{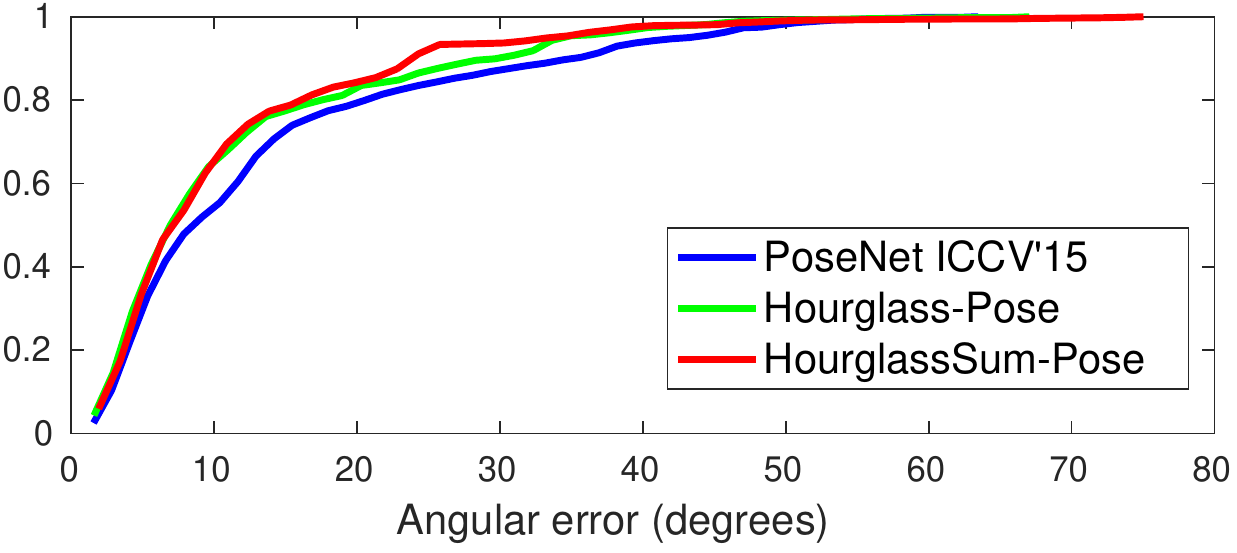}%
		\includegraphics[width=.5\textwidth]{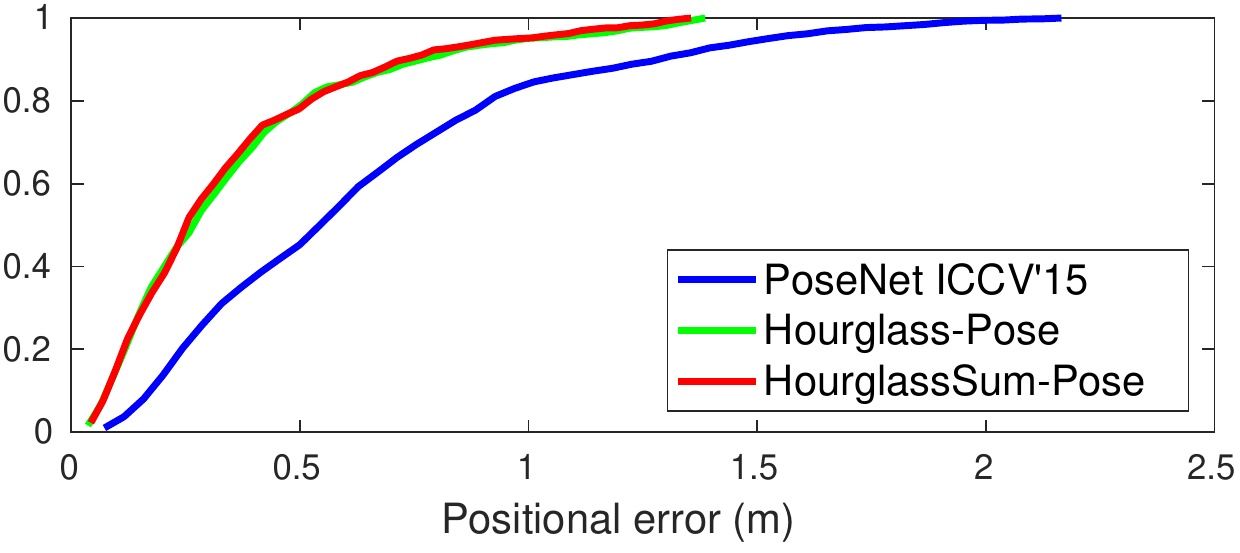}
		\caption{Pumpkin}
	\end{subfigure}%
	~
	\begin{subfigure}[t]{.5\textwidth}
		\centering
		\includegraphics[width=.5\textwidth]{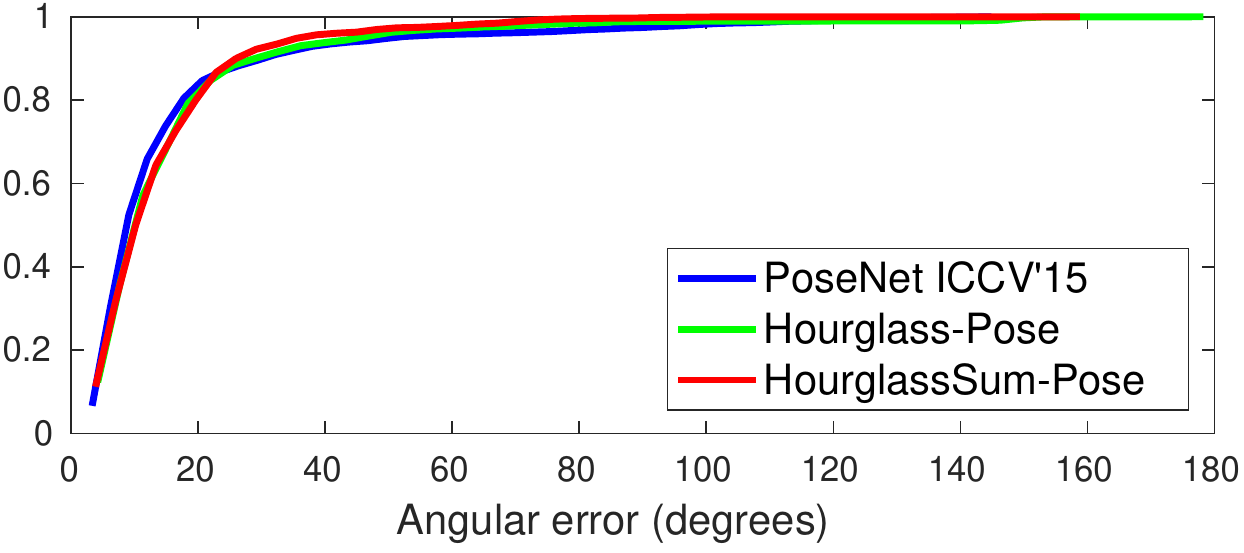}%
		\includegraphics[width=.5\textwidth]{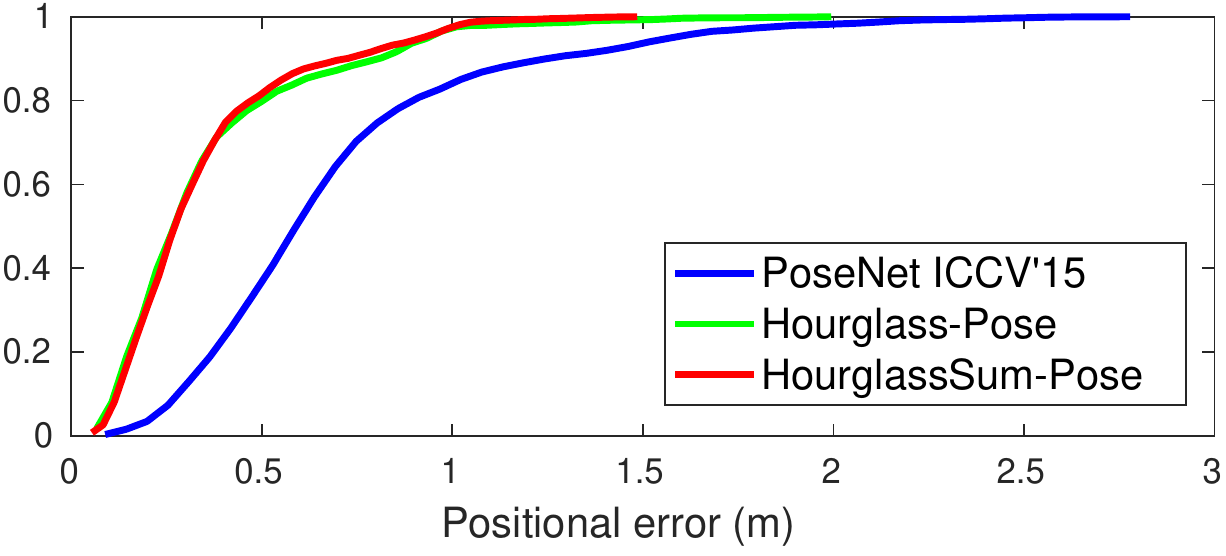}
		\caption{Red Kitchen}
	\end{subfigure}
	~
	\begin{subfigure}[t]{.5\textwidth}
		\centering
		\includegraphics[width=.5\textwidth]{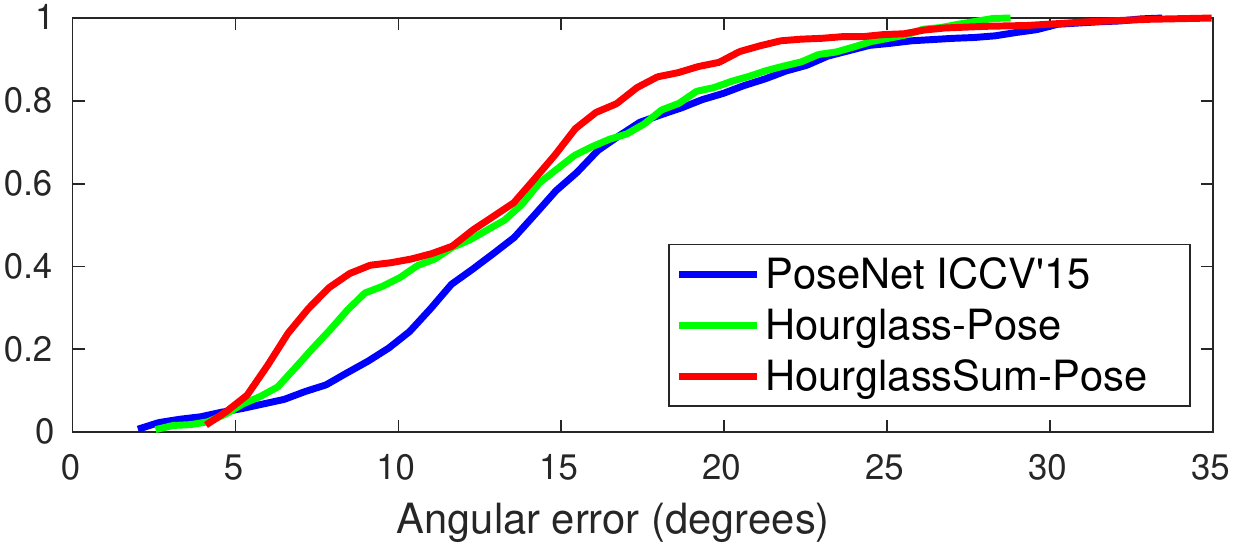}%
		\includegraphics[width=.5\textwidth]{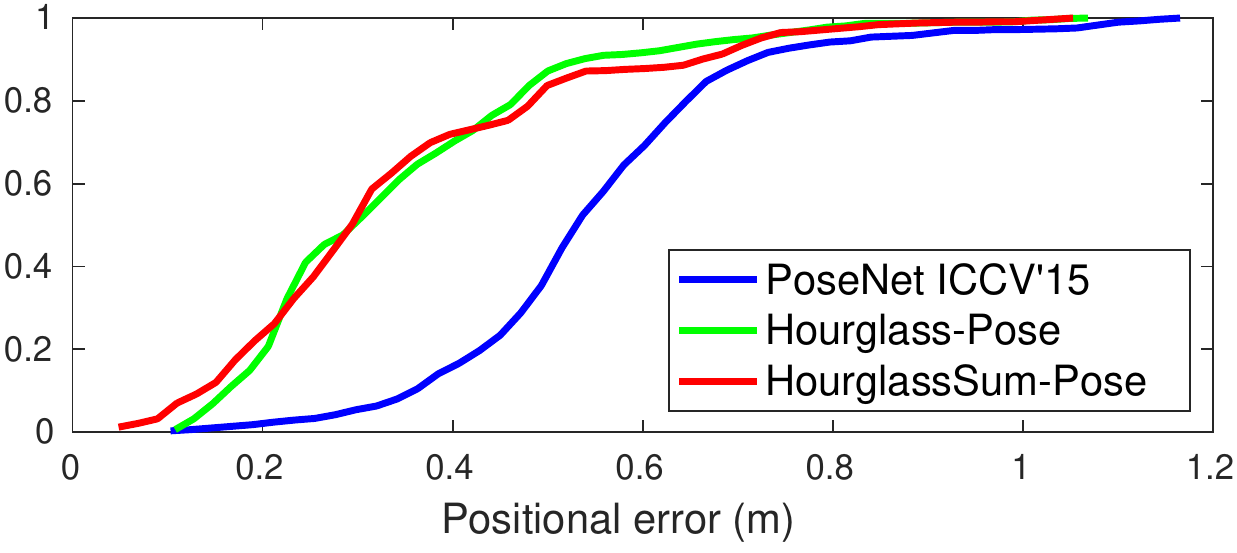}
		\caption{Stairs}
	\end{subfigure}
\caption{Localization performance of the proposed hourglass-based network architectures (Hourglass-Pose and HourglassSum-Pose) presented as a cumulative histogram (normalized) of errors for all categories of 7-Scenes dataset. One of the important conclusion is that both architectures can significantly improve the accuracy of estimations camera location clearly outperforming state-of-the-art method (PoseNet). HourglassSum-Pose achieves better orientation performance in 5 cases to compare to Hourglass-Pose architecture. }\label{lbl:fig_cumulative_hist}
\end{figure*}

For a more detailed comparison, we plot a family of cumulative histogram curves for all scenes of the evaluation dataset illustrated in Fig.~\ref{lbl:fig_cumulative_hist}. We note that both hourglass architectures outperforms PoseNet method on translation accuracy by a factor of 1.5 to 2.3 in all test scenes. Besides that, HourglassSum-Pose substantially improves orientation accuracy. The only exception is 'Office' and 'Red Kitchen' scenes where performance of HourglassSum-Pose is on par with PoseNet.
%they provide substantially more accurate pose than PoseNet in all test scenes. 

\begin{figure}[t!]
	\centering
	\begin{subfigure}[t]{.5\textwidth}
		\centering
		\includegraphics[width=.5\textwidth]{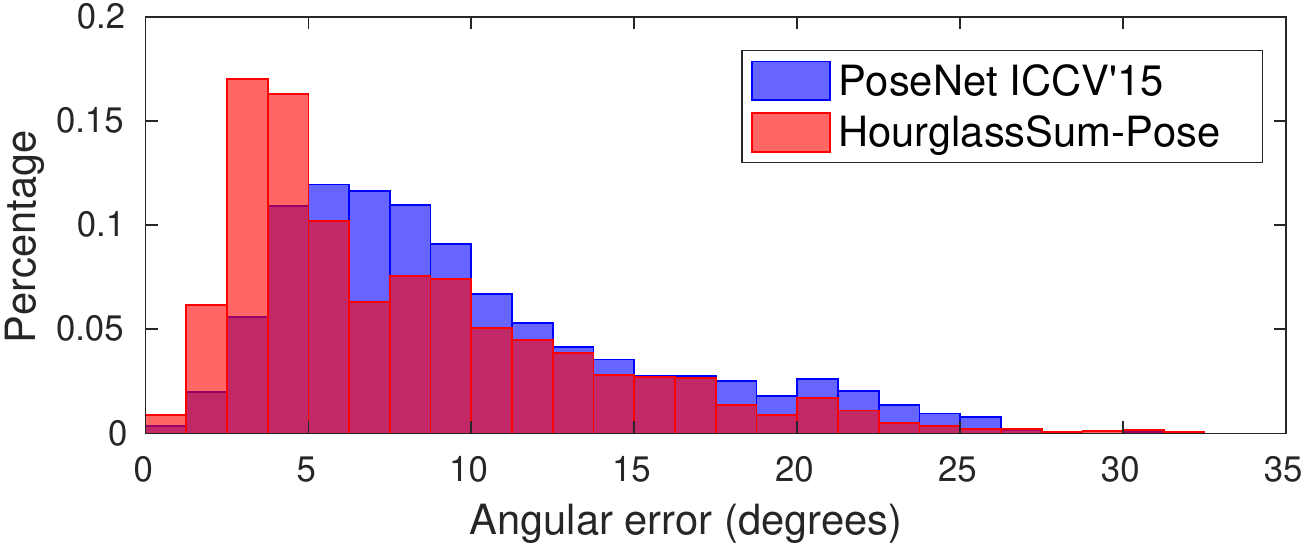}%
		\includegraphics[width=.5\textwidth]{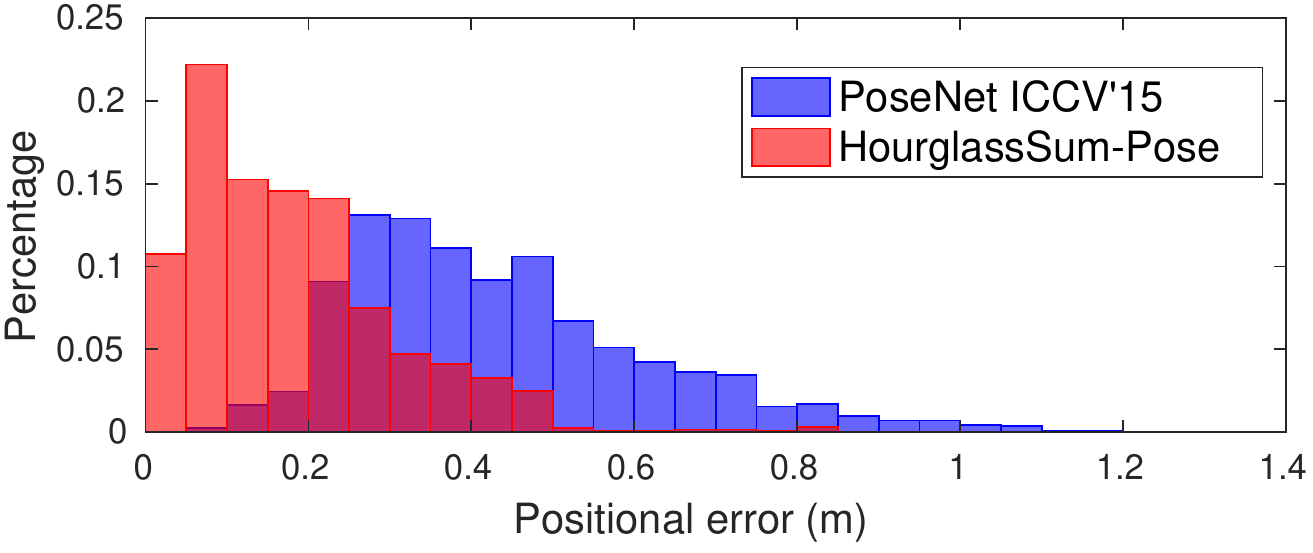}
		\caption{Chess}\label{lbl:fig_hist_chess}
	\end{subfigure}
	~
	\begin{subfigure}[t]{.5\textwidth}
		\centering
		\includegraphics[width=.5\textwidth]{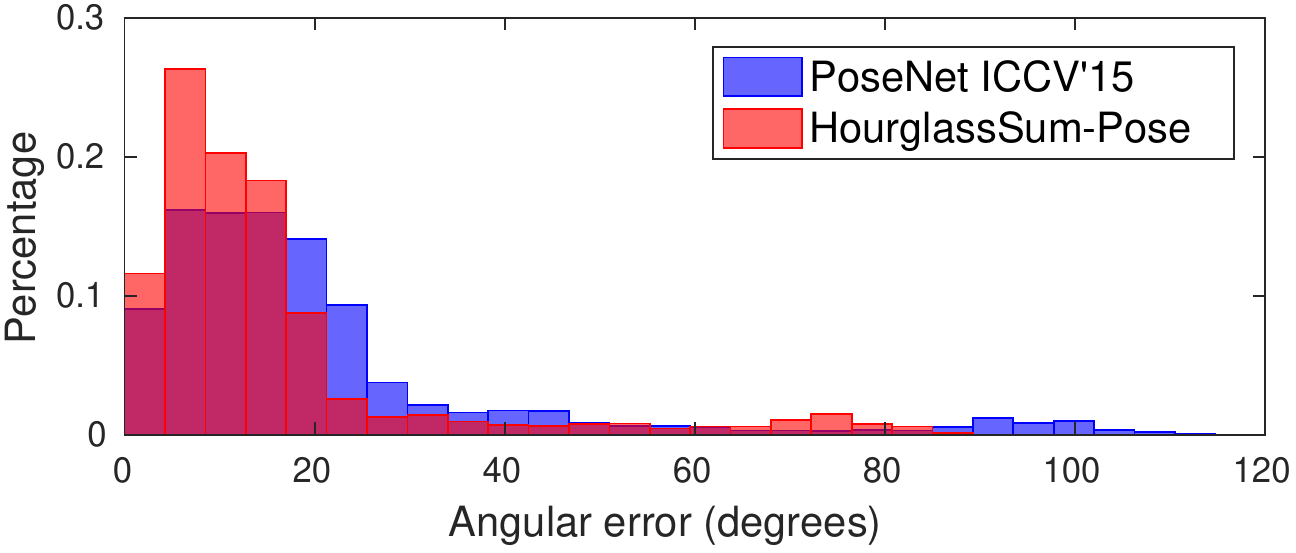}%
		\includegraphics[width=.5\textwidth]{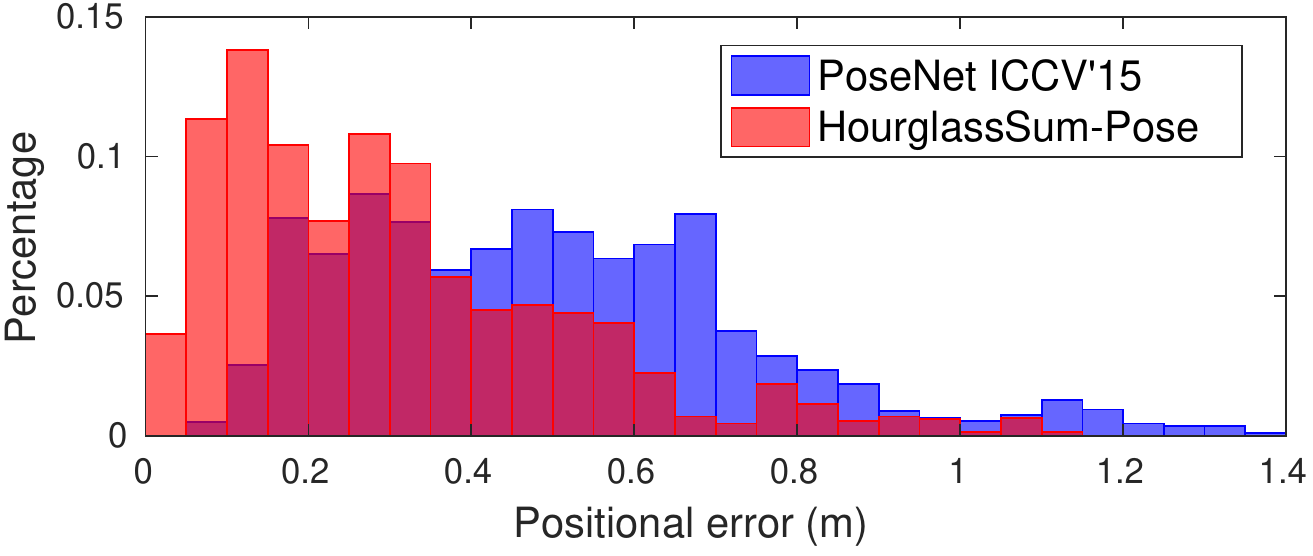}
		\caption{Stairs}\label{lbl:fig_hist_stairs}
	\end{subfigure}
\caption{Histogram of orientation (left) and translation (right) errors of two approaches (PoseNet and HourglassSum-Pose) for the two entire scenes ('Chess' and 'Fire') of the evaluation dataset. It is clearly seen that an hourglass-architecture-based method performs consistently better than PoseNet.}\label{lbl:fig_histograms}
\end{figure}

Figure~\ref{lbl:fig_histograms} shows histograms of localization accuracy for both orientation (left) and position (right) for the two entire test scenes of the evaluation dataset. It is interesting to see that more than 60\% of camera pose estimations produced by HourglassSum-Pose are within 20 cm in 'Chess' scene, while for PoseNet this quotient is equal to 5\%. Remarkably, HourglassSum-Pose is able to improve accuracy even for such an ambiguous and challenging scene like 'Stairs' exhibiting many repetitive structures (See Fig.~\ref{lbl:fig_hist_stairs}). The presented results verify that an hourglass neural architecture is an efficient and promising approach for image-based localization.

\section{Conclusion}\label{lbl:sec_conclusion}
In this paper, we have presented an end-to-end trainable CNN-based approach for image-based localization. One of the key aspect of this work is applying encoder-decoder (hourglass) architecture consisting of a chain of convolutional and up-convolutional layers for estimating 6-DoF camera pose. Furthermore, we propose to use direct connections forwarding feature maps from early residual layers of the model directly to the later up-convolutional layers improving the accuracy. We studied two hourglass models and showed that they significantly outperform other state-of-the-art CNN-based image-based localization approaches.

%We propose an approach for adding context to a state-of-the-art object detection framework, and demonstrate its effectiveness on benchmark datasets. While we expect many improvements in finding more efficient and effective ways to combine the features from the encoder and decoder, our model still achieves good performance outperforming state-of-the-art relocalization results on 7-Scenes dataset. 

{\small
\bibliographystyle{ieee}
\bibliography{egbib}
}

\end{document}